\documentclass[11pt,a4paper]{article}

\usepackage[hyperref]{acl2020}

\usepackage{times}
\usepackage{latexsym}
\usepackage{graphicx}
\usepackage[inline]{enumitem}
\usepackage{array}
\usepackage{multirow}
\usepackage{caption}
\usepackage{subcaption}
\newcommand{\model}[1]{{\small{#1}}}
\usepackage{amsmath}
\usepackage{relsize}
\usepackage{arydshln}
\usepackage{booktabs}
\usepackage{xcolor}

\usepackage[utf8]{inputenc}

\usepackage{times}
\usepackage{latexsym}

\usepackage{microtype}

\usepackage{url}
\aclfinalcopy

\title{Multi-agent Communication meets Natural Language:\\Synergies between Functional and Structural Language Learning}

\author{
Angeliki Lazaridou\thanks{~All authors contributed equally.} , Anna Potapenko\footnotemark[1] , Olivier Tieleman\footnotemark[1]\\
DeepMind, London, UK\\
\texttt{\{angeliki,apotapenko,tieleman\}@google.com}}

\date{}

\begin{document}
\maketitle
\begin{abstract}
We present a method for combining multi-agent communication and traditional data-driven approaches to natural language learning, with an end goal of teaching agents to communicate with humans in natural language.  Our starting point is a language model that has been trained on generic, not task-specific language data. We then place this model in a multi-agent self-play environment that generates task-specific rewards used to adapt or modulate the model, turning it into a task-conditional language model. We introduce a new way for combining the two types of learning based on the idea of reranking language model samples, and show that this method outperforms others in communicating with humans in a visual referential communication task. Finally, we present a taxonomy of different types of language drift that can occur alongside a set of measures to detect them.
\end{abstract}

\section{Introduction}
\label{sec:intro}

In this work, we aim at making agents communicate with humans in natural language. Our starting point is a language model that has been trained on generic, not task-specific language data. We then place this model in a multi-agent communication environment that generates task-specific rewards, which are used to adapt or modulate the model, making it task-conditional. We thus propose to decompose the problem of learning language use into two components: learning ``what'' to say based on a given situation, and learning ``how'' to say it.  The ``what'' is the essence of communication that underlies our intentions and is chosen by maximizing any given utility, making it a \textit{functional}, utility-driven process. On the other hand, the ``how'' is a surface realization of our intentions, i.e., the words we use to communicate this ``what'' successfully.  This factorization into \textit{content planning} (here, ``what'') and \textit{surface realization} (here, ``how'') moves us away from end-to-end neural generation systems and is in line with traditional methods of natural language generation~\cite{Reiter:Dale:1997}. More importantly, it enables us to bring together two different strands of research: traditional data-driven natural language learning and multi-agent communication.

Traditional approaches to natural language learning~\cite{Kneser:Ney:1995,Mikolov:etal:2010,Sutskever:etal:2014, Vinyals:Le:2015, Radford:etal:2019} are based on inferring \textit{structural} properties of language from text corpora, often in a passive regime, dissociated from communication. While this type of learning is great for learning general statistical associations between symbols (e.g., adjectives come before nouns) and even inferring semantic relations, it ignores the functional aspects of communication, i.e., the fact that people use words to coordinate with others and make things happen in the world~\cite{Wittgenstein:1953,Austin:1975,Clark:1996}.

On the other hand, multi-agent communication research~\cite{Foerster:etal:2016,Lazaridou:etal:2017,Havrylov:Titov:2017,Evtimova:etal:2017,Lee:etal:2019} puts communication at the heart of agents' (language) learning. Implemented within a multi-agent reinforcement learning setup, agents start \textit{tabula rasa} and form communication protocols that maximize task rewards. While this purely utilitarian framework results in agents that successfully learn to solve the task by creating a communication protocol, these emergent communication protocols do not bear core properties of  natural language.  \citet{Chaabouni:etal:2019} show that protocols found through emergent communication, unlike natural language, do not conform to Zipf’s Law of Abbreviation;~\citet{Kottur:etal:2017} find that communication protocols do not follow compositionality patterns of natural language, and~\citet{Lazaridou:etal:2018} find emerged protocols to be sensitive to different experimental conditions. This growing set of alarming results on emergent communication raises doubts about the use of this type of \textit{functional} learning as a viable alternative to language learning.

Concluding that neither approach on its own is adequate for learning language use, we propose a  method for combining the best of both worlds. Generic language data can be used effectively as a good prior model of language,  encapsulating its intrinsic \textit{structural} properties, i.e., are only used for the ``how'' in the form of generic language models. Conversely, multi-agent interactions, that provide rewards specific to the task of interest, now only need to be used for the \textit{functional} learning of language use, i.e., learning the ``what''.\footnote{About the terminology: by `traditional data-driven natural language learning', we mean language modelling of the next-word-prediction variety. This type of learning does not involve any \textit{use} of the language or other context, and as such only focuses on word statistics. Since the structure of the language is a large part of those statistics, and the role of the generic language models in our proposed combined systems is to provide structural knowledge of language, we also use the term `structural learning'. We contrast this with the purely usage-driven, reward-based learning of the type seen in emergent communication research. Since the function, rather than the structure or statistics, is the only thing that matters for such a learner, we also use the term `functional learning'.}

The contributions of this paper are as follows. First, we propose a general research program of language learning  that combines two learning signals coming from multi-agent communication and traditional data-driven natural language learning techniques. We present a concrete study in the context of a referential communication game (see Section~\ref{sec:framing}) between a speaker and a listener, where the traditional data-driven language learning takes the form of image captioning, and the functional learning takes the form of agent self-play (see Section~\ref{sec:setup}). We then present a new approach for combining the two learning signals, i.e., reward-learned rerankers (see Section~\ref{section:methods}), and compare this to existing approaches using a human study (see Section~\ref{sec:ref_success}). We discuss shortcomings of this program with respect to different types of \textit{language drift} that can occur, and introduce a number of automatic measures to detect them (see Section~\ref{sec:language_drift}). Finally, we show how such a program under oracle rewards can be a viable approach moving towards learning language use from human rewards (see Section~\ref{sec:oracle}).

\section{Research framing}
\label{sec:framing}

Our research can be framed in the following scenario. An agent needs to perform a \textit{functional communication task} in a natural language (in this work, English). However, examples of linguistic communication  about this functional task are not available - the only natural language data that can be used consist of examples of \textit{generic natural language}, which are not grounded in the functional task. Recasting the task as a multi-agent language game provides a way to obtain a \textit{reward} that judges whether an utterance elicited the correct behaviour by a listener.

\subsection{Experimental setup}

In this work, we instantiate the research in the following way: the \textit{functional task} is a visual referential communication game for a target image in the context of a distractor, the \textit{reward} is based on success in referential communication where a listener is tasked to pick the correct image within  distractors guided by the speaker's description, and the \textit{generic natural language data} are captioning data.

\paragraph{Visual referential communication game.} There are two players, the speaker and the listener. The speaker sees a target object and needs to communicate an utterance about it in the context of distractors; both target and distractors are represented as images. The listener is presented with the same set of images, but without the knowledge of which is the target, and needs to identify the target image relying on the utterance being communicated by the speaker. The utterance takes the form of sequences of word-like units. If the listener's choice is correct they both receive a positive reward, else they receive the same negative reward.\footnote{The task we consider is essentially \textit{discriminative image captioning}~\cite{Vedantam:etal:2017,Dai:etal:2017,Andreas:Klein:2016}. Here we are using it as a placeholder of a communication task to illustrate our general framework. Thus, we are not incorporating any explicit bias in the model about this particular task. The only task-specific information we use is communicated via the reward.}

\begin{figure}
    \begin{subfigure}[b]{.2\textwidth}
        \includegraphics[scale=0.26]{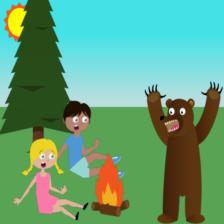}
    \end{subfigure}%
\begin{subfigure}[b]{.4\textwidth}
{\footnotesize{
        Jenny is scared of the bear
        \smallskip
        
        Mike is scared of the bear
        \smallskip
        
        Jenny and Mike sit by a fire
        \smallskip
       
        Jenny and Mike are sitting
        
        A bear is scaring mike and jenny
}}
   \end{subfigure}
    \caption{Example image and ground-truth captions from the Abstract Scenes dataset used in this study.}
    \label{fig:exampleimages}
\end{figure}

\paragraph{Dataset and referential splits.} For playing the visual referential communication game, we use a multi-modal dataset, the Abstract Scenes~\cite{Zitnick:Parikh:2013} which contains 10k synthetic images accompanied with descriptive captions (on average 6 per image) (see Figure~\ref{fig:exampleimages}).\footnote{Other multi-modal datasets like MSCOCO~\cite{Lin:etal:2014} or Flickr~\cite{Thomee:etal:2016}, while providing complex naturalistic images, often have a repetitive set of captions, highlighting one particular aspect of the scene and suffer from a human reporting bias~\cite{Misra:etal:2016}.
By using Abstract Scenes, we have left certain visual challenges out of the scope of the work, obtained cleaner multi-modal associations between words and objects, and focused on the language use for referential communication.}
The captions typically refer to diverse aspects of the scene (characters and actions), providing a rich and challenging environment for an agent to evolve the captioning skills for successful communication.  In our experiments, we split the dataset into 80/10/10  for train/validation/test sets. We use the test images to create two referential splits, i.e., \textit{easy} and \textit{difficult}, as a function of the similarity between the target and distractor images. Each split contains $1000$ pairs of a target and a distractor.

\begin{table}[t]
\small
    \centering
    \begin{tabular}{ccc}
    \toprule
    \multicolumn{1}{c}{speaker (human)} &
    \multicolumn{1}{c}{easy}  & 
    \multicolumn{1}{c}{difficult}   \\\midrule
  \multirow{1}{*}{\model{random}} & 0.92 & 0.81 \\\midrule
  \multirow{1}{*}{\model{discriminative}}&  1.0 & 0.97\\
   \bottomrule
    \end{tabular}
    \caption{Accuracy performance of a human listener with a human speaker producing either \model{random} or \model{discriminative} caption on the easy and difficult splits.}
    \label{tab:gold_caption}
\end{table}

\paragraph{Human performance and setup validation.}
\label{sec:task_validation}
In order to assess the difficulty of the task in the presence of the particular data (images and captions) we perform a human study in the reference game with a human speaker and a human listener, where the human speaker can only communicate one of the existing captions of the target image. We perform the human study under two conditions. In the first condition, the human speaker has only access to the ground-truth captions and does not have access to the distractor image, thus has to pick a \model{random} caption. This corresponds to the \textit{perfect} structural knowledge of English but \textit{no} knowledge of the functional task and it is the human upper-bound of a captioning system performance on this task. In the second condition, the speaker has access to both the ground-truth captions and the distractor image, thus is able to pick a \model{discriminative} caption to communicate. For each condition, we collect 50 rounds of games and present results in Table~\ref{tab:gold_caption}. We see that the task-specific condition outperforms the first condition, indicating that in our current setup there is enough space to improve upon models based on structural-only learning (i.e., captioning models). 
Moreover, the good performance of \model{discriminative} caption speaker demonstrates that (in principle) the captioning data can be used in a successful communication with a human for this task.

\section{Multi-agent communication setup}
\label{sec:setup}

\subsection{Speaker}
\label{sec:setup_speaker}
The speaker is the primary learner in this research, aiming at creating a model that is able to use natural language in a communicative scenario, and consists of standard visual and language modules. To convert images to embeddings $u$, we use a pre-trained ResNet~\cite{He:etal:2016}  (parametrized by $\theta^{resnet}$) and feed its last layer output into a one-layer MLP (parametrized by $\theta^{MLP}_{S}$). 
To generate a message $m$, we use a one-layer LSTM~\cite{Hochreiter:Schmidhuber:1997} (parametrized by $\theta^{LSTM}_{S}$), adding embeddings $u$ at each time step as additional context. 
Section~\ref{section:methods} presents different speaker models consisting of these modules.

We also design two oracle speakers (with no weights) that have direct access to ground-truth captions of images at test time.  The \model{random} caption speaker outputs one of the ground-truth captions for the target image at random. Since this speaker is not aware of the functional goal, their performance will indicate whether having only good grounded language skills is enough for communication success in our setup.  We also build an oracle speaker that is task-aware; \model{discriminative} caption speaker uses a simple word-overlap heuristic to pick the target's caption that has the least word overlap with any of the distractor's captions (the score is normalized by the captions' length excluding stop-words).

\subsection{Listener}
\label{sec:listener_setup} Throughout the experiments, we need a way to estimate performance on the functional communication task, either for evaluation or to provide rewards during training acting as a scaffolding to learn the speaker model. Ideally, this performance signal should be provided by a human who is interacting online with the speaker agent. While we do so for evaluation reasons,  for training we approximate this quantity with a learned component, an agent listener. 

To convert images to embeddings $u$, we use the same pre-trained ResNet as for the speaker and feed its last layer output into a one-layer MLP (parametrized by $\theta^{MLP}_{L}$). Following that, the listener uses an LSTM  (parametrized by $\theta^{LSTM}_{L}$) to embed the utterance $m$ received by  speaker, creating embedding $v$. Finally, the listener picks the image with the highest dot-product similarity between the embedded message $v$ and the embeddings $u_t$ and $u_d$ for target and distractor. Since we know which image candidate is the intended referent, we cast this problem as supervised learning and update the listener's weights $\theta_L=\{\theta^{MLP}_{L}$, $\theta^{LSTM}_{L}\}$ optimizing cross-entropy. Finally, the listener assigns reward 1 to the speaker if they identified the correct image, else reward -1. 

We consider two different setups: a \model{joint} listener, which is trained together with the speaker, as commonly done in the emergent communication literature, and a \model{fixed} listener that is pre-trained to perform best-response to the oracle \model{discriminative} caption speaker and stays fixed throughout the learning of the speakers with the sole use of providing them rewards. We expect the latter setup to be less prone to language drift issues due to the grounding of the \model{discriminative} caption speaker to language data. thus potentially resulting in better communication with human listeners. We also use the \model{fixed} listener for evaluation of all speakers.

\section{Methods for learning language use}
\label{section:methods}

We describe ways to estimate the speaker's generative model~$p_{\theta_{S}}(m|u, t)$ for message~$m$, conditioned on target and distractor embeddings~$u=[u_t, u_d]$ and target image index~$t\in\{0,1\}$.

\subsection{Functional-only learning}
\label{sec:functional}
This type of learning language use is identical to experiments commonly conducted in the literature of \model{emergent communication}~\cite{Lazaridou:etal:2017, Havrylov:Titov:2017, Bouchacourt:Baroni:2018, Evtimova:etal:2017, Graesser:etal:2019}, i.e., the speaker learns to emit communication utterances $m$ in order to maximize the communication task reward (see Section~\ref{sec:listener_setup} for a discussion on how this reward is computed). Concretely, the weights $\theta_{S}=\{\theta^{MLP}_{S}$, $\theta^{LSTM}_{S}\}$ of the speaker policy~$\pi_{\theta_{S}}(m|u, t)$ are updated via the REINFORCE update rule~\cite{Williams:1992} using rewards $r^{L}$ provided by the listener, i.e., we optimize $L^{functional}=-r^{L}(m, u, t) \mathlarger{\sum}_{i=1}^I \log p_{\theta^{LSTM}_{S}}(m^i|m^{<i}, u)$, where  $u=[u_t;u_d]$, $m^i \in V$, vocabulary size $|V|=100$, and message length $I=10$.\footnote{In all experiments using REINFORCE we add an entropy regularization term to the loss.} Note, that while this type of learning results in a language that is maximally functionally correct for the given task reward, this language is not natural language, i.e., the symbols are not grounded to natural language.

\vspace*{-2pt} 
\subsection{Structural-only learning}
\label{sec:structural}
This type of learning ignores the functional aspect of communication and communicates utterances that reflect intrinsic structural properties of language, i.e., that are fluent, grammatical and related to the target. Here, we used paired data in the form $\langle u, c\rangle$, where $u$ is a visual embedding and $c$ is the associated caption, and learn an \model{image captioning} model. The speaker's parameters  $\theta_{S}=\{\theta^{MLP}_{S}$, $\theta^{LSTM}_{S}\}$ are optimized using cross-entropy, i.e.,
$L^{structural}=-\mathlarger{\sum}_{i=1}^I \log p_{\theta^{LSTM}_{S}}(c^i|c^{<i}, u)$, 
where $u=u_t$, $c^i \in V$, $|V|=2685$ (the vocabulary size) and $I=25$, i.e., the longest caption in the dataset.
We approximate the speaker model $p_{\theta_{S}}(m|u, t)$ with the captioning one, which ignores distractor, thus the communication task. 
We construct two speakers with different decoding schemes: \model{greedy} uses greedy decoding, while \model{sample} picks the highest probability message among $k=20$ stochastic samples (temperature $\tau$=2.0).

\subsection{Structural and functional learning}
\label{sec:combination}
We now describe several ways in which both types of learning are used to learn language use. In all cases, we equip the speaker with a base image captioning model similar to the one presented in Section~\ref{sec:structural}, which is used to calculate $p_{\theta^{LSTM}_{S}}(c^i|c^{<i}, u_t)$.
The functional part is learned via the REINFORCE update rule optimizing the task reward (i.e., listener's accuracy in the referential task). However, speakers differ in how they parametrize $p_{\theta_{S}}(m|u, t)$ and whether the task reward is used to update the weights $\{\theta^{MLP}_{S}, \theta^{LSTM}_{S}\}$ of the base captioning model.

\subsubsection{Reward finetuning}
\label{sec:finetuning}
The simplest approach is to \textit{first} use existing pre-trained components for which we have available corpora in order to learn the statistical properties of language, and \textit{then} steer the language use to be functionally appropriate using \model{reward finetuning} for the given task. We use paired data in the form $\langle u, c\rangle$ to learn the weights $\theta_{S}=\{\theta^{MLP}_{S}$, $\theta^{LSTM}_{S}\}$  of a base image captioning model following Section~\ref{sec:structural}, and then we perform functional learning by using the listener's reward to optimize the weights $\theta_{S}$ as in Section~\ref{sec:functional}. While this method is conceptually simple, it becomes challenging when the task requires extending the conditioning part of the base model. Here, we need to change the conditioning of the base captioning model from $u=u_t$ to $u=[u_t;u_d]$,  to allow conditioning on the distractor. 
Since this is not trivial (the base image captioning model has been learned by conditioning only on one image embedding), we keep the conditioning~$u=u_t$ also during finetuning with REINFORCE.
Thus, similar to the \model{image captioning} model, we approximate $p_{\theta_{S}}(m|u, t)$  with $p_{\theta^{LSTM}_{S}}(m|u_t)$. However,  unlike \model{image captioning}, the information about distractors flows into the policy, since
the weights $\theta_{S}$ are optimized using the listener's reward  which considers distractors. 

Since the gradients from optimizing the functional task are sent all the way into the base captioning model, this causes catastrophic forgetting of the \textit{core knowledge} of language, leading to \textit{language drift}. Thus, we use a language regularizer term in the form of Kullback-Leibler divergence between pre-trained and fine-tuned language modeling distributions~\cite{Havrylov:Titov:2017}.

\subsubsection{Multi-task learning}
\label{sec:multitask}
An alternative is to conduct both types of learning (i.e., image captioning and functional learning) at the same time~\cite{Lazaridou:etal:2017, Lee:etal:2019}. This takes the form of multi-task learning optimizing $\lambda_{f}L^{functional}+\lambda_{s}L^{structural}$, where $\lambda_{f}=1$. Like in \model{reward finetuning}, the gradients of the reward learning flow into the weights of a base captioning model, leaving us with questions about a trade-off between task success and quality of language.  Therefore, we introduce two variants of this model depending on the importance of the language component, i.e., one variant with $\lambda_{s}=0.1$ and  a language-regularized one with $\lambda_{s}=1$.

\subsubsection{Reward-learned rerankers}
\label{sec:rerankers}
Finally, we introduce a new way of learning language use in the multi-agent communication setup. 
As before, we train the core language capabilities of a speaker using the image captioning task objective, but after this pre-training phase, the weights of this model are frozen. The functional part is then viewed as \textit{learning to use} this general knowledge of language grounded in images.
This is operationalized as learning to \textit{rerank} samples obtained from the captioning model optimizing the listener's reward. 
The action space of this speaker are sentences, as opposed to words used commonly in the literature of emergent communication. 
We emphasize that by leveraging the idea of reranking, we are able to take a \textit{task-unconditional} model, i.e., a captioning model that only conditions on the target, and extend its conditioning turning it into a \textit{task-conditional} model, i.e., a discriminative captioning model that conditions also on the distractor.

Below we consider two concrete reranker models. 
In both cases, the message generation happens in two steps. First, we sample $|S|=20$ candidates from the pre-trained and fixed image captioning model~$p_{\theta_{S}^{LSTM}}(m|u_t)$. Then, we pick the best sample~$s$ using a {task-conditional} reranking score~$p(s|u, t)$. 
The reranking score can be viewed as a new policy~$\pi_{\theta_S}(s|u, t)$ that operates in the space of samples $S$ drawn from the task-unconditional model.
This policy introduces an additional set of trainable parameters~$\theta_S^{rerank}$ that are learned with REINFORCE.
Thus, the full set of weights for this speaker is $\theta_{S}=\{\theta^{MLP}_{S}$, $\theta^{LSTM}_{S}, \theta_S^{rerank}\}$. 
Crucially, the two learning signals, i.e., \textit{structural} and \textit{functional}, affect different set of weights, i.e.,  $\{\theta^{MLP}_{S}$, $\theta^{LSTM}_{S}\}$ and $\theta_S^{rerank}$ respectively, allowing   submodules to specialize.

\textbf{Product of experts reranker.} 
In this model we parametrize the policy as a product of experts (PoE):
$\pi_{\theta_S}(s|u, t) \propto p(s|u, t)^{\lambda_{f}} p(s|u_t)^{\lambda_{s}}$,
where~$u=[u_t; u_d]$ and $\lambda_{f}=1$. 
The second term is the image captioning message probability, \textit{re-normalized} over the samples space, thus  bringing general language knowledge grounded in images.
The first term adjusts for the task specifics. 
To model that, we re-embed the samples using transformed bag-of-words, thus the trainable parameters of the reranker~$\theta_S^{rerank}$ are word embeddings and additional MLP weights. We combine target and distractor embeddings into a single vector and compute the dot-product similarity between this vector and each of the bag-of-words representations of samples. Finally, these scores are passed through a softmax layer to obtain~$p(s|u, t)$. We introduce two variants of the model, one with $\lambda_{s}=0$ and a language-regularized one with $\lambda_{s}=1$.

\textbf{Noisy channel reranker.} 
Following Bayes rule, we factorize the speaker's policy as follows:
$\pi_{\theta_S}(s|u, t) \propto p(t|s, u) p(s| u)$, where~$u=[u_t; u_d]$.
We omit the distractor vector~$u_d$ in the conditioning of the prior, arriving to~$p(s| u_t)$ from the PoE reranker above.
The crucial difference is that the first term now represents the speaker's approximation of the listener's behaviour.
As before, we represent samples with the transformed bag-of-words, but then compute their dot-product similarities with each image separately and normalize with softmax across the images to obtain the probability of the target~$p(t|s, u)$.  This reranker model is closely related to pragmatic speakers in Rational Speech Act (RSA) framework~\cite{Andreas:Klein:2016,Monroe:Potts:2015,Cohn-Gordon:etal:2018,Fried:etal:2018}. However, while the RSA model assumes a given and fixed listener model, here we are learning the model of the listener that the speaker is using by optimizing end-to-end  the listener's reward. Thus, when doing multi-agent communication using the \model{noisy channel} model, there exist two components that produce probability distributions of the same type $p(t|s, u)$; one belongs to the listener, thus the speaker has no access to it (e.g., this listener in the future could be a human sending rewards), while the other belongs to the speaker corresponding to their model of the listener.

\section{Speakers trained jointly with listeners}
\label{sec:ref_success}

Table~\ref{tab:combo_table} presents referential success when speakers are \textit{trained} with rewards from a \model{joint} listener, i.e., a listener being learned jointly with the speaker. 

We conduct three different evaluations: at \textit{test} time we play against the \model{fixed} listener, \textit{human} listeners and the \model{joint} listener the speaker was trained with. While \model{fixed} listener is the same for all speakers, the \model{joint}  listener is speaker-specific. We report results on two splits: for the \textit{easy} and \textit{difficult} split we report referential success of the \model{joint} listener, and for the latter split, we also report results of the \model{fixed} and \model{human} listener. 

To compute referential success using \model{human} listeners, we collect 400 annotations for each speaker model. To avoid annotators adapting to model-specific strategies, we group predictions of similar models and collect annotations in three sessions (one for each group), during which we present annotators with predictions from a model sampled from that group.\footnote{Group 1: \model{image captioning (greedy/sample)}, \model{noisy channel}, \model{PoE}). Group 2: \model{multi-task},  \model{reward finetuning}. Group 3: \model{random}, \model{discriminative}, \model{PoE} and \model{noisy channel} with ground-truth captions.}

\begin{table}[t!]
\small
\hspace{-0.4cm}
\begin{tabular}{p{0.18\textwidth}>{\color{gray}}cccc}
\toprule
  \multicolumn{1}{l}{} & 
  \multicolumn{1}{c}{\color{gray}Easy split} &  \multicolumn{3}{c}{Difficult split} \\
  \multicolumn{1}{l}{}      & \multicolumn{1}{c}{\color{gray}joint} &  
  \multicolumn{1}{c}{joint} &  
  \multicolumn{1}{c}{fixed} &
\multicolumn{1}{c}{human}\\
  \midrule
 \multicolumn{5}{l}{\bf  Functional-only learning} \\
 \model{emergent} (§\ref{sec:functional}) & 0.99 & 0.98 & -& 0.5\\
 \midrule
 
 \multicolumn{5}{l}{\bf  Structural-only learning} \\
  \multicolumn{5}{l}{\model{image captioning}  (§\ref{sec:structural})} \\
 \model{~~~sample}      & 0.92 & 0.78    &    0.77  & 0.77 \\  
\model{~~~greedy}  & 0.91 & 0.77    &    0.73  & 0.78 \\
\midrule
\multicolumn{5}{l}{\bf  Structural \& functional learning} \\
\multicolumn{5}{l}{\it Gradients from reward affect base captioning model} \\
\multicolumn{5}{l}{\model{reward finetuning }(§\ref{sec:finetuning})}\\ 
\model{~~no KL-term} & 0.95 & 0.82    &    0.63    &    0.62  \\
\model{~~with KL-term} & 0.93  & 0.79    &    0.77    &    0.69\\ 
\multicolumn{5}{l}{\model{multi-task} learning (§\ref{sec:multitask})}\\
~~$\lambda_{s}=0.1$  & 0.98 & 0.94    &    0.71    &    0.71    \\ 
~~$\lambda_{s}=1$ &    0.96 &     0.90    &    0.69    &    0.69   \\ 
\multicolumn{5}{l}{\it Reranking (§\ref{sec:rerankers}), base captioning model unchanged} \\
\model{PoE}, $\lambda_{s}=0$  & 0.99 &  0.92    &    0.81   & 0.81 \\ 
\model{PoE}, $\lambda_{s}=1$   & 0.98 &  0.91    &    0.83  & 0.78  \\  
\model{noisy channel} & 0.96 & 0.83    &    0.84  & 0.86*  \\ 
 \midrule
\multicolumn{5}{l}{\bf Utilizing ground-truth captions from the dataset} \\
\multicolumn{5}{l}{\it Oracle speakers, no weights learned~(§\ref{sec:setup_speaker})} \\
\model{random}& 0.87 &  0.74    &    0.72  & 0.81\\  
\model{discriminative} & 0.87 &  0.73    &    0.82  & 0.87* \\ 
\multicolumn{5}{l}{\it Reranking (§\ref{sec:rerankers}) ground-truth captions}\\
\model{PoE}~(§\ref{sec:rerankers})  & 0.95 &  0.88    &    0.85  & 0.93* \\
\model{noisy channel}~(§\ref{sec:rerankers})  & 0.95 & 0.78    &    0.83  & 0.88* \\
\bottomrule
\\
\end{tabular}
\caption{Referential success of speakers (by rows) \textit{trained} with \model{joint} listener and then \textit{tested} with \model{joint}, \model{fixed} and \model{human} listener (by columns).*~indicates significance over the \model{image captioning (greedy)} when \textit{tested} with humans ($p<0.005$, bootstraping test). }
\label{tab:combo_table}
\end{table}

\subsection{Referential success of joint listeners}
All models perform quite similarly in the \textit{easy} split, whereas we observe larger gaps in the difficult split. In terms of \model{joint} accuracy results in the difficult split, \model{reward finetuning} has the lowest performance among models that are optimizing rewards, perhaps due to its large action space (i.e., the vocabulary size $|V|=2685$), making it a hard RL exploration problem. \model{multi-task}, despite having the same action space performs better, probably due to the captioning objective being optimized concurrently facilitating the learning dynamics. Finally, the best results in both splits are obtained by the \model{emergent communication} model, that achieves near perfect performance. We believe this is the case since this speaker is the least constrained of all, since we can think of all other speakers (i.e., the ones that combine both types of learning) as being regularized towards producing natural language.

\subsection{Referential success of human listeners}
Somewhat alarmingly, we observe the \model{joint} performance is not predictive of the \model{human}'s one across the board, hinting to issues regarding \textit{pragmatic} drift (we will further discuss this in Section~\ref{sec:language_drift}). In the most extreme case, while the \model{emergent communication} speaker achieved the highest results when playing against a listener jointly learned with the speaker, this comes in the expense of human performance: functional learning alone results in maximally uninterpretable protocols, and as such humans are at random when playing against such a model.

Speakers that combine both types of learning achieve good human performance,
with reward-learned reranker models, i.e., \model{noisy channel} and \model{PoE} being the best. In their case, they outperform the \model{image captioning} baselines, even approaching the \model{discriminative} oracle speaker based on ground-truth captions. This indicates their effectiveness in extending the conditioning of the underlying \model{image captioning} to the distractor image with the reward coming from the \model{listener}, turning like this the base image-captioning model into a task-specific referential captioning model. Moreover, when giving the rerankers a perfect captioning model in the form of ground-truth captions of target images,  performance of  \model{noisy channel} and \model{PoE} surpass the oracles' (see last two columns of Table~\ref{tab:combo_table}); as the community improves the base language models, we should expect this to also result in net improvement in the reranker models.

Finally, we also observe that the \model{fixed} grounded listener is significantly predictive of the \model{human} performance ($p<0.005$, t-test).\footnote{All t-tests are conducted between two distributions of scores dichotomized on human performance.} This is encouraging, since as we will show in Section~\ref{sec:oracle}, we can use this listener as a fixed model that provides rewards to the speaker model.

\section{Language drift and how to detect it}
\label{sec:language_drift}

We show that the multi-agent communication framework is prone to language drift~\cite{Lee:etal:2019}, i.e., when protocols diverge from human language. We present a taxonomy of different types that occur in this framework, alongside a set of  automatic measures to detect it.

\begin{table}[t!]
\footnotesize
\begin{tabular}{p{1.9cm} l}
\toprule
\hspace{0.1cm}Target Image &
\hspace{1cm} Distractor Image\\
\includegraphics[scale=0.25]{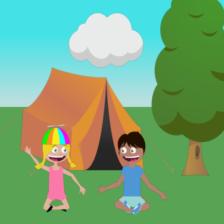} &  \hspace{1cm}\includegraphics[scale=0.25]{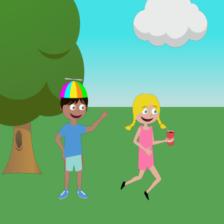}\\\midrule
\multicolumn{2}{l}{\bf  Structural-only learning} \\
\multicolumn{2}{l}{image captioning~(§\ref{sec:structural})} \\
\model{~~sample}  &  \textbf{jenny} is wearing a hat \\
\model{~~greedy} &  \underline{mike}  is wearing a hat \\[-0.7em] \\\cline{1-2} \\[-0.7em]
\multicolumn{2}{l}{\bf  Structural and functional learning} \\
\multicolumn{2}{l}{\it Gradients from reward affect base captioning model} \\
\multicolumn{2}{l}{reward finetuning~(§\ref{sec:finetuning})} \\
\model{~no KL-term} & it is camping \textbf{camping} [...] camping \\
\model{~with KL-term} & mike is sitting \underline{on} the tent \\
\multicolumn{2}{l}{\model{multi-task} learning (§\ref{sec:multitask})} \\
~~$\lambda_{s}=0.1$ &  mike is jenny on \underline{the the} tent\\
~~$\lambda_{s}=1$ & mike is sitting on the ground \\
 \\[-0.7em]\cdashline{1-2} \\[-1em]
\multicolumn{2}{l}{\it Reranking (§\ref{sec:rerankers}), base captioning model unchanged} \\
\model{PoE}, $\lambda_{s}=0$  &  the tent is in the \underline{tree} \\
\model{PoE}, $\lambda_{s}=1$ &  mike and jenny are sitting \textbf{on the ground} \\
\model{noisy channel}  & jenny is wearing a \textbf{funny hat} \\
\bottomrule
\\
\end{tabular}
\caption{Examples of generated messages. We underline  \underline{wrong} and bold \textbf{correct} cases of language usage.}
\label{tab:examples}
\end{table}

\begin{table}[t]
\small
\hspace{-0.4cm}
\begin{tabular}{p{0.12\textwidth}>{}cccc}
\toprule
  \multicolumn{1}{l}{}      & 
  \multicolumn{1}{c}{log~p(m)} &  
  \multicolumn{1}{c}{log~p(m$|$i)} &  
  \multicolumn{1}{c}{1-gram} &
  \multicolumn{1}{c}{3-gram}\\
\midrule
 
 \multicolumn{5}{l}{\bf  Structural-only learning} \\
  \multicolumn{5}{l}{\model{image captioning} (§\ref{sec:structural})} \\

 \model{~~sample} &  -8.71    &    -7.77    &    0.81    &     0.37   \\  
\model{~~greedy}  &     -8.63    &    -7.72    &    0.73*    &      0.30*  \\
\midrule
\multicolumn{5}{l}{\bf  Structural \& functional learning} \\
\multicolumn{5}{l}{\it Gradients from reward affect base captioning model} \\
\multicolumn{5}{l}{\model{reward finetuning }(§\ref{sec:finetuning})}\\ 
\model{~~no KL-term} & -442.00    &    -279.55    &    0.33    &    0.00  \\
\model{~~with KL-term} & -11.75    &    -10.78    &    0.70*    &     0.22* \\ 
\multicolumn{5}{l}{\model{multi-task} learning (§\ref{sec:multitask})}\\
~~$\lambda_{s}=0.1$  &  -18.08*    &    -19.67    &    0.78*    &  0.18*    \\ 
~~$\lambda_{s}=1$    &   -10.68*    &    -10.64    &    0.63* &    0.18   \\ 
\hdashline \\[-1em] 
\multicolumn{5}{l}{\it Reranking (§\ref{sec:rerankers}), base captioning model unchanged} \\
\model{PoE}, $\lambda_{s}=0$   &  -10.18*    &    -8.79    &    0.71*    &    0.24*  \\ 
\model{PoE}, $\lambda_{s}=1$   &  -8.95    &    -7.94    &    0.78*    &   0.30*  \\  
\model{noisy~channel}  &  -10.02    &    -8.59    &    0.76    &  0.30  \\ 
\bottomrule
\\
\end{tabular}
\caption{Language drift measures, lower scores mean higher drift. *indicates that the measure was significantly predictive of the human listener performance on the referential task (p$<$0.005, t-test).}
\label{tab:drift_measures}
\end{table}

\subsection{Structural drift: Definition and measures} 
The most basic type of drift that manifests in the emergent communication setup relates to the core structural properties of the generated language, i.e.,  its fluency and grammaticality with respect to \textit{natural language} (this is also referred to by \citet{Lee:etal:2019} as ``syntactic''). Looking at Table~\ref{tab:examples}, a clear example of this type of drift happens when models update the base captioning model. \model{reward finetuning} (no KL-term) does not produce at all grammatical sentences, while \model{multi-task} ($\lambda_s=0.1$) appears to suffer less, only occasionally producing slightly ungrammatical sentences by repeating consecutive words. We term this \textit{structural drift} and we quantify it as the log probability of the generated message under a pre-trained unconditional language model (column \model{log~p(m)} in Table~\ref{tab:drift_measures}). 

\subsection{Semantic drift: Definition and measures} The second type of drift is the  \textit{semantic} drift. This relates to whether the generated message is grounded with regards to the target object, i.e., its adequacy with respect to the literal semantics of the target  (this is also referenced by \citet{Lee:etal:2019} as ``semantic''). We have qualitatively observed instances of this type of drift in the \model{PoE}, which occasionally shifts the semantics of words, e.g., using the word \textit{tree} to refer to \textit{ground} as seen in Table~\ref{tab:examples}. To measure it, we use a pre-trained image-conditional language model and compute the target-image conditional log probability of the generated message (column \model{log~p(m$|$i)} in Table~\ref{tab:drift_measures}). 

These two log probability-based measures do not assume access to language data for the target objects, and as such can be computed from general unconditional and domain-specific conditional language models. In this particular case though, since we also have access to language data for the target images (i.e., captions in English), and assuming that these data describe everything that is true about the target, we can use simple n-gram statistics as proxies of semantic drift (i.e., in this case \model{1-gram} word overlap ignoring stop word and \model{3-gram} word overlap between the ground-truth captions and the speaker-generated message). Moreover, all these measures do not take into account the specific communication task the speaker has to perform, i.e., our measures do not consider any information about the distractor object, making them easily adaptable to other tasks. 

\subsection{Structural and semantic drift results}
In Table~\ref{tab:drift_measures} we report performance of different models under these automatic measures. The structural score \model{log~p(m)} reflects the qualitative observations made from Table~\ref{tab:examples}, i.e.,  \model{multi-task} and \model{reward finetuning}, have the highest structural drift, with the latter performing significantly worse than all the models. In contrast, the reranker models that do not update the base captioning model, i.e., \model{PoE} and \model{noisy channel}, perform the best on the semantic score by construction; both models directly incorporate in their models a component associated with the semantic score (i.e., the samples taken from the image-conditional model alongside the associated probabilities).  Moreover, they also perform well on all other measures,  indicating their robustness against language drift. Finally, all the model-specific language regularizers (KL-term for \model{reward finetining}, $\lambda_{s}=1$ for \model{multi-task} and $\lambda_{s}=1$ for \model{PoE}) we introduced were effective in limiting both types of language drift (as also seen in Table~\ref{tab:examples}).

\subsection{Pragmatic drift}
Finally, we identify a novel type of drift, i.e., \textit{pragmatic} drift, which relates to the divergence between a human's interpretation of the message from the interpretation a speaker will assume. Unfortunately, this type of drift is perhaps the most difficult to capture in an automatic way as it is \textit{task specific} and requires access to the exact interpretation that the human would ascribe to the message. 
As a proxy of pragmatic drift, we use the difference between the agent- and human-listener referential success; if the \model{joint} referential success is higher than the human's one, then the  speaker assumes an interpretation of the message that is different from the human's one, resulting in lower human performance.  An extreme example of this drift manifests when the \model{joint} listener achieves almost perfect referential success whereas a human listener is at random, as in the case of \model{emergent communication}. However, in this case the messages are maximally uninterpretable with the lowest possible performance in both structural and semantic scores. 

Hence, a natural question to ask is \textit{to what degree (if at all possible) pragmatic drift can manifest in the absence of the other two types of language drift.} 
Or, put differently, does the emergent communication for learning language use hide any other pathological behaviour for models that do not suffer a lot from structural and semantic drift, as in the case of \model{PoE} and \model{noisy channel}? 
To study this, we create a setup where \model{PoE} is guaranteed to have a perfect knowledge of (grounded) language. Namely, it uses the reward to rerank \textit{ground-truth captions} associated with the target image (note, our dataset provides up to five captions per image).
Moreover, we perform several ablations where we allow the updating of different parameters in the speaker's and listener's model by unfreezing components.%

\begin{table}[h]
\footnotesize
\begin{tabular}{lccc} 
\toprule
 \multicolumn{1}{l}{Weights learned with RL} & 
 \multicolumn{1}{c}{\model{joint}}  &
 \multicolumn{1}{c}{\model{human}} & 
 \multicolumn{1}{c}{$\Delta$}
 \\\midrule
reranker  & 0.88 & 0.92 & -0.04\\
reranker + speaker ResNet  & 0.92 & 0.90 & +0.02\\
reranker + both agent ResNets  & 0.96 & 0.88 & +0.08\\
\bottomrule
 \end{tabular}
\caption{Referential success for \model{PoE} with gold captions when updating different components during training with \model{joint} listener.}
\label{tab:drift}
\end{table}

Table~\ref{tab:drift} presents the results of the \model{joint} and \model{human} referential success. The main finding is that by increasing the number of components that get updated using the \model{joint} reward, the margin between the referential success of the two types of listeners increases. Despite the fact that the  speaker is using human language that is perfectly fluent and accurate with respect to the target image (since the reranker operates on captions associated with the target image), while the \model{joint} listener is able to communicate with the agent speaker, the human listener achieves significantly lower performance. 

In one test example, the speaker said \textit{Mike has a hat}, which was equally true for both images making the human pick at random. So, how could the listener pick correctly? The speaker had reached a pact with the listener that the interpretation of this message will be something beyond what the phrase means (e.g., \textit{Mike has a yellow hat} or the intensity of the pixels in the target image is lower). Since speaker and listener learn together, they co-adapt, forming conventions (or conceptual pacts~\cite{Brennan:Clark:1996}) that differ from humans', even in the presence of \textit{fluent} and \textit{grounded} language.

\section{Speakers trained using fixed listener}
\label{sec:oracle}

In the previous section we showed that learning a speaker using a learned reward module as a scaffolding (i.e., the \model{joint} listener) can lead to pragmatic drift. In this section, we use a grounded reward as scaffolding. In the absence of a human listener to provide rewards for learning, we use the oracle \model{fixed} listener, which was found in Section~\ref{sec:ref_success} to be predictive of human referential success. It is pre-trained, stays fixed and just provides rewards for training the speaker. As speakers, we use the models that scored the highest in Table~\ref{tab:combo_table} and retrain them against \model{fixed}. Table~\ref{tab:oracle listener} presents the results of referential success against \model{fixed} and \model{human} listeners.
Using a grounded reward results in better performance for the weaker models. The small gap between the rerankers in the two experimental setups points that using a learned reward module (\model{joint}) holds promise, despite the different types of language drift. Moreover, we show that our models for learning language can be used against fixed reward models, potentially learning directly from human rewards~\cite{Ziegler:etal:2019b}.

\begin{table}[t!]
\footnotesize
\centering
\begin{tabular}{lc c} 
\toprule
 \multicolumn{1}{l}{Model}  & 
 \multicolumn{1}{c}{\model{fixed}}  &
 \multicolumn{1}{c}{\model{human}} \\\midrule
\model{reward finetuning, with KL-term}  & 0.81  & 0.75 \\
\model{multi-task} learning, $\lambda_s = 0.1$  & 0.80  & 0.68 \\
\model{PoE}, $\lambda_s = 0$   & 0.93 & 0.86 \\  
\model{noisy channel}  &  0.88 &  0.87 \\
\bottomrule
\end{tabular}
\caption{Referential success when training speakers with the \model{fixed} listener.}
\label{tab:oracle listener}
\end{table}

\section{Discussion and Limitations}

We presented a method for teaching agents to communicate with humans in natural language, by \textbf{combining two learning signals coming from multi-agent communication and traditional data-driven natural language learning techniques}, which adds on recent efforts of blending emergent communication with natural language~\cite{Lowe:etal:2020, Lu:etal:2020}.

Self-play between speakers and listeners can result in \textbf{language drift, the most severe of which being pragmatic drift}. 
Since speakers and listeners are learning concurrently, they can co-adapt to pair-specific policies that deviate from the policies that humans learn. This pathological behaviour of self-play is not specific to language and extends to other policies~\cite{Carroll:etal:2019}. 

Finally, we introduced the \textbf{reward-learned reranker approach which alleviates language drift and achieves the highest human performance}, by constraining the functional learning to happen on the level of utterances generated by a pre-trained language model. However, since the functional signal is not currently influencing the sampling from the language model, this will lead to poor performance when using more general language models with weaker conditioning (e.g. GPT-2~\cite{Radford:etal:2019}) whose samples potentially do not fit the functional context. 
Moving towards integrating our findings into more realistic applications of self-play, e.g., user simulation in dialogue~\cite{Schatzmann:etal:2006, Pararth:etal:2008}, these shortcomings need to be addressed. 

\section*{Acknowledgements}
We thank Kris Cao, Laura Rimell, Chris Dyer, Phil Blunsom, Aida Nematzadeh and the DeepMind Language Team for useful feedback, Susie Young and Adam Liska for help with  annotations.

\bibliography{citations}
\bibliographystyle{acl_natbib}

\appendix

\section{Appendices}
\label{sec:appendix}

%
%



\paragraph{Hyperparameters.}
The following tables represent our choice of hyper-parameters in the speaker and listener agents. 
Hyperparameters in Table~\ref{tab:hparams_shared} where chosen in the image captioning task using the validation set.
Hyperparameters in Table~\ref{tab:hparams_specific} where chosen in the referential task using the validation set.

\begin{table}[h!]
\footnotesize
\begin{tabular}{c|c|c} 
 \multicolumn{1}{c}{Agent}  & 
 \multicolumn{1}{c}{Hyperparameter}  &
 \multicolumn{1}{c}{Value} \\ \hline\hline
 listener  & LSTM hidden size  & 512 \\  
 speaker  & LSTM hidden size  & 512 \\ 
 listener  & visual embeddings size  & 512 \\  
 speaker  & visual embeddings size  & 1024 \\
\end{tabular}
\caption{Settings shared across all experiments.}
\label{tab:hparams_shared}
\end{table}

\begin{table}[h!]
\footnotesize
\begin{tabular}{c|c|c} 
 \multicolumn{1}{c}{Model}  & 
 \multicolumn{1}{c}{Hyperparameter}  &
 \multicolumn{1}{c}{Value} \\ \hline\hline
 \model{fine-tuning++}  & KL regulatization  & 0.1 \\ 
 \model{multi-task}  & structural weight  & 0.1 \\ 
 \model{PoE}  & structural weight  & 0 \\ 
  \model{noisy channel} / \model{PoE} & number of samples  & 20 \\
  \model{noisy channel} / \model{PoE} & message embedding size  & 1024 \\
  \model{noisy channel} /  \model{PoE} & entropy regularization  & 0.1 \\

\end{tabular}
\caption{Settings for particular speakers.}
\label{tab:hparams_specific}
\end{table}

\paragraph{ResNet module.}
We use ResNet-50~\cite{He:etal:2016} pre-trained on ImageNet.
For \model{image captioning} and also for models that use the pre-trained captioning model (i.e. \model{reward finetuning}, \model{PoE} and \model{noisy channel}) we back-propagate gradients into the ResNet module. However, in all rerankers we freeze the ResNet  during reward optimization. Moreover, we also keep the ResNet fixed in the jointly learned listener to prevent additional drift, however we back-propagate when we pre-train the fixed listener, grounded though the \model{discriminative} caption speaker.

\end{document}